\documentclass[11pt]{article}
\usepackage{eacl2017}
\usepackage{times}
\usepackage{url}
\usepackage{latexsym}
\usepackage{amsmath,graphicx, tikz, ulem}
\usepackage{microtype}
\usepackage{multirow, multicol}
\usepackage[utf8]{inputenc}
\usepackage{cleveref}
\usepackage{phaistos}
\usepackage{color}
\usepackage{soul}

\crefname{section}{§}{§§}
\Crefname{section}{§}{§§}

\newcommand\todo[1]{}

\usepackage{tabularx}
\usepackage{multirow}

\newcolumntype{C}{>{\centering\arraybackslash}X}
\newcolumntype{L}{>{\raggedright\arraybackslash}X}

\eaclfinalcopy 
\def\eaclpaperid{352} 

\title{Towards speech-to-text translation without speech recognition}

\author{
{\it Sameer Bansal$^{1}$, Herman Kamper$^{2}$, Adam Lopez$^{1}$, Sharon Goldwater$^1$}\\
$^{1}$School of Informatics, University of Edinburgh \\
$^{2}$Toyota Technological Institute at Chicago, USA \\
{\tt\small \{sameer.bansal, sgwater, alopez\}@inf.ed.ac.uk, kamperh@gmail.com} \\
}

\date{23-Nov-2017}

\usepackage{xcolor}
\definecolor{orange}{HTML}{FF6600}

\newcommand{\herman}[1]{}
\newcommand{\sameer}[1]{}
\newcommand{\sg}[1]{}

\begin{document}
\maketitle
\begin{abstract}

We explore the problem of translating speech to text in low-resource scenarios where neither automatic speech recognition (ASR) nor machine translation (MT) are available, but we have training data in the form of audio paired with text translations. We present the first system for this problem applied to a realistic multi-speaker dataset, the CALLHOME Spanish-English speech translation corpus. Our approach uses unsupervised term discovery (UTD) to cluster repeated patterns in the audio, creating a \textit{pseudotext}, which we pair with translations to create a parallel text and train a simple bag-of-words MT model. We identify the challenges faced by the system, finding that the difficulty of cross-speaker UTD results in low recall, but that our system is still able to correctly translate some content words in test data.

\end{abstract}

\section{Introduction} 
\label{sec:intro}

Typical speech-to-text translation systems pipeline automatic speech recognition (ASR) and machine translation (MT)~\cite{waibel+fugen_ieee08}.
But high-quality ASR requires hundreds of hours of transcribed audio,
while high-quality MT requires millions of words of parallel text---resources available for only a tiny fraction of the world's estimated 7,000 languages~\cite{besacier2014}. Nevertheless, there are important low-resource settings in which even limited speech translation would be of immense value: documentation of endangered languages, which often have no writing system~\cite{besacier+etal_slt06,AlanBlackStuff}; and crisis response, for which text applications have proven useful~\cite{munro2010}, but only help literate populations. In these settings, target translations may be available. For example, ad hoc translations may be collected in support of relief operations. Can we do anything at all with this data?

In this exploratory study, we present a speech-to-text translation system that learns directly from source audio and target text pairs, and does not require intermediate ASR or MT.
Our work complements several lines of related recent work. For example,
 \newcite{duong2015attentional} and \newcite{antonios+chiang+duong_EMNLP2016} presented models that align audio to translated text, but neither used these models to try to translate new utterances (in fact, the latter model cannot make such predictions). \newcite{berard+etal_nipsworkshop16} did develop a direct speech to translation system, but presented results only on a corpus of synthetic audio with a small number of speakers. Finally, Adams et al.\ \shortcite{adams+etal_interspeech16,adams+etal_emnlp16} targeted the same low-resource speech-to-translation task, but instead of working with audio, they started from word or phoneme lattices. In principle these could be produced in an unsupervised or minimally-supervised way, but in practice they used supervised ASR/phone recognition. Additionally, their evaluation focused on phone error rate rather than translation.
In contrast to these approaches, our method can make translation predictions for audio input not seen during training, and we evaluate it on real multi-speaker speech data.


Our simple system (\textsection\ref{sec:system}) builds on unsupervised speech processing~\cite{versteegh+etal_interspeech15,lee+etal_tacl15,kamper+etal_taslp16}, and in particular on \textit{unsupervised term discovery} (UTD), which creates hard clusters of repeated word-like units in raw speech~\cite{park2008unsupervised,jansen2011efficient}. The clusters do not account for all of the audio, but we can use them to simulate a partial, noisy transcription, or \textit{pseudotext}, which we pair with translations to learn a bag-of-words translation model. We test our system on the CALLHOME Spanish-English speech translation corpus~\cite{post2013improved}, a noisy multi-speaker corpus of telephone calls in a variety of Spanish dialects (\textsection\ref{sec:dataset}).
Using the Spanish speech as the source and English text translations as the target,
we identify several challenges in the use of UTD, including low coverage of audio and difficulty in cross-speaker clustering (\textsection\ref{sec:utd_analysis}). Despite these difficulties, we demonstrate that the system learns to translate some content words (\textsection\ref{sec:experiments}).

\section{From unsupervised term discovery to direct speech-to-text translation}\label{sec:system}





\begin{figure}[!t]
    {\centering
        \centerline{\includegraphics[width=0.95\linewidth]{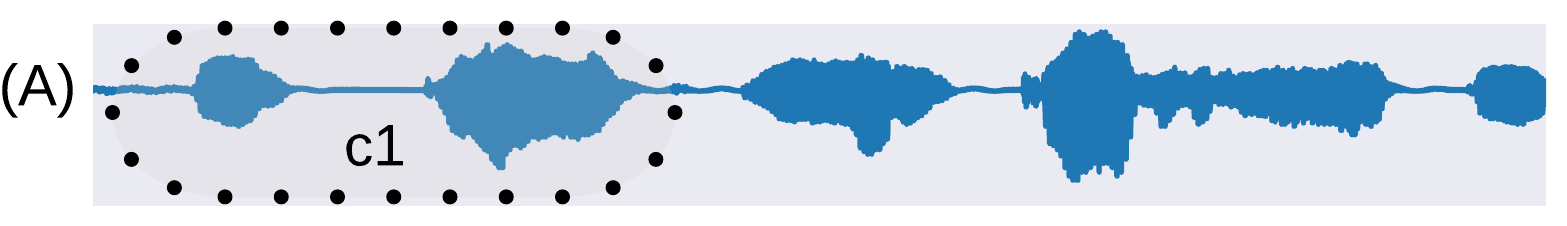}}
    }
    \vspace*{-0.3\baselineskip}
    \hspace*{0.7\baselineskip}
    {\small
        \begin{tabularx}{\linewidth}{ll}
    		{Spanish:} & \uline{{\color{gray} sí} pues} {\color{gray} y} {\color{gray} el} carrito \\
    		{English:} & yes well {\color{gray} and} {\color{gray} the} car \\
    		{Pseudotext:} & c1 \\
		\end{tabularx}
	}
	\vspace*{-0.3\baselineskip}
   	{\centering
    	\centerline{\includegraphics[width=0.95\linewidth]{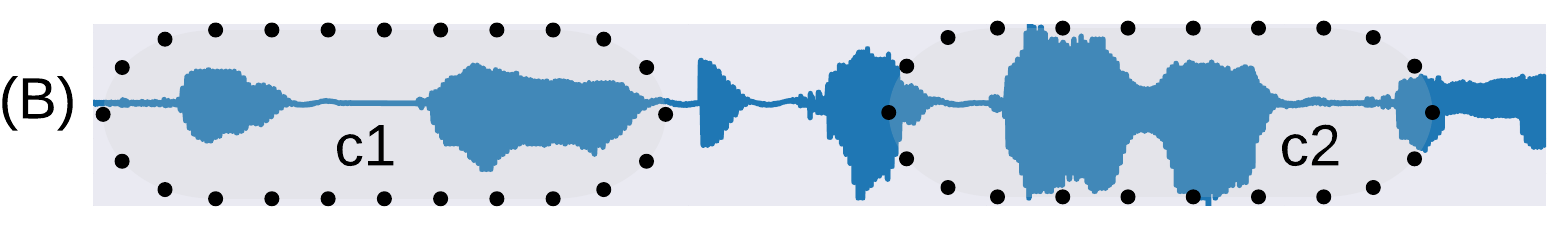}}
	}
    \vspace*{-0.4\baselineskip}
    \hspace*{0.7\baselineskip}
    {\small
        \begin{tabularx}{\linewidth}{ll}
    		{Spanish:} & \uline{{\color{gray} sí} pues} {\color{gray} y} qu\uline{é tal vas {\color{gray} co}}{\color{gray}n} \\
			{English}: & yes well {\color{gray} and} hows {\color{gray} it} going \\
			{Pseudotext}: & c1, c2 \\
		\end{tabularx}
	}

	\vspace*{0.2\baselineskip}
	{\centering
	    \centerline{\includegraphics[width=0.95\linewidth]{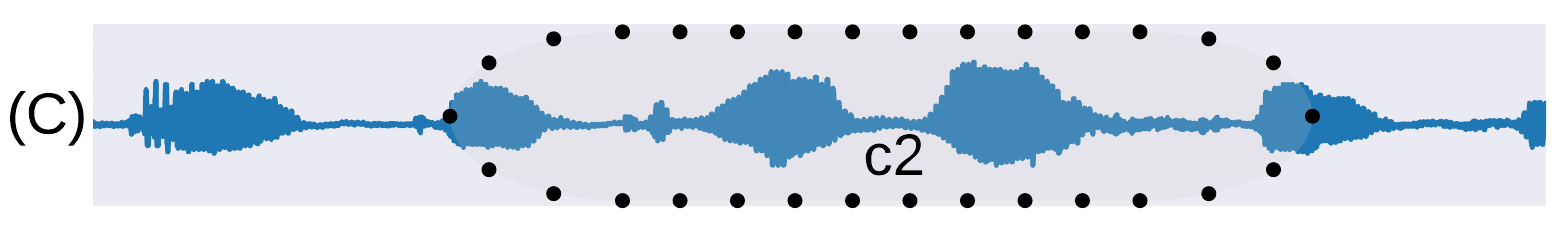}}
    }
    \vspace*{-0.1\baselineskip}
    \hspace*{0.7\baselineskip}
    {\small
        \begin{tabularx}{\linewidth}{ll}
    			{Spanish:} & {\color{gray} est}\uline{{\color{gray}e} trabajo} {\color{gray} y} {\color{gray} se}  \\
    			{English}: & {\color{gray} this} work \\
    			{Pseudotext}: & c2 \\
    	\end{tabularx}
	}

	\vspace*{-0.1\baselineskip}
   	{\centering
        \centerline{\includegraphics[width=0.95\linewidth]{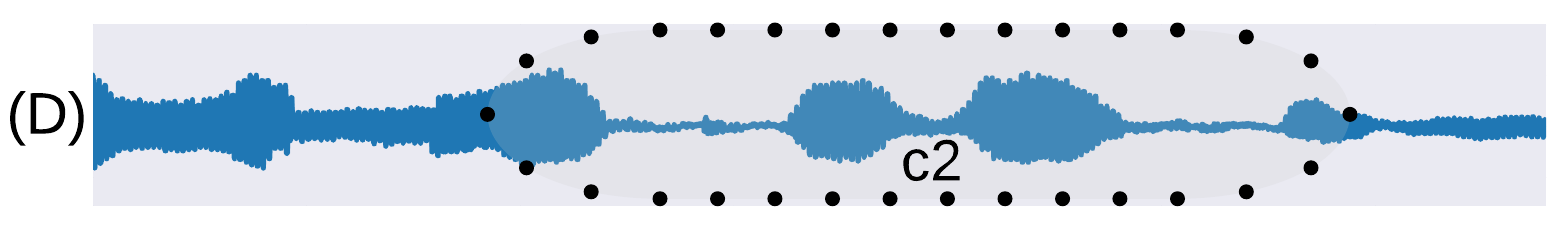}}
	}
    \vspace*{-0.1\baselineskip}
    \hspace*{0.7\baselineskip}
    {\small
        \begin{tabularx}{\linewidth}{ll}
    			{Spanish:} & llama {\color{gray} d}\uline{{\color{gray}el} trabajo} {\color{gray} mi} \\
    			{English}: & call {\color{gray} him} {\color{gray} from} work \\
    			{Pseudotext}: & c2 \\
		\end{tabularx}
	}
    \vspace*{-0.3\baselineskip}
    \caption{Example utterances from our data, showing UTD matches, corresponding pseudotext, and English translation. For clarity, we also show Spanish transcripts with the approximate alignment of each pseudoterm underlined, though these transcripts are unavailable to our system. Stopwords (in gray) are ignored in our evaluations. These examples illustrate the difficulties of UTD: it does not match the full audio, and it incorrectly clusters part of utterance B with a good pair in C and D.}
    \label{fig:utd_out}
    \vspace*{-0.7\baselineskip}
\end{figure}

For UTD we use the Zero Resource Toolkit (ZRTools; Jansen and Van Durme, 2011).\footnote{\texttt{https://github.com/arenjansen/ZRTools}} \nocite{jansen2011efficient} ZRTools uses dynamic time warping (DTW) to discover pairs of acoustically similar audio segments, and then
uses graph clustering on overlapping pairs to form a hard clustering of the discovered segments.
Replacing each discovered segment with its unique cluster label, or \textit{pseudoterm}, gives us a partial, noisy transcription, or pseudotext (Fig.~\ref{fig:utd_out}).

In creating a translation model from this data, we face a difficulty that does not arise in the parallel texts that are normally used to train translation models: the pseudotext does not represent all of the source words, since the discovered segments do not cover the full audio (Fig.~\ref{fig:utd_out}). Hence we must not assume that our MT model can completely recover the translation of a test sentence. In these conditions, the language modeling and ordering assumptions of most MT models are unwarranted, so we instead use a simple bag-of-words translation model based only on co-occurrence:
IBM Model 1~\cite{brown+1993} with a Dirichlet prior over translation distributions, as learned by {\tt fast\_align}~\cite{dyer+2013+fastalign}.\footnote{We disable diagonal preference to simulate Model 1.} In particular, for each pseudoterm, we learn a translation distribution over possible target words. To translate a pseudoterm in test data, we simply return its highest-probability translation (or translations, as discussed in \textsection\ref{sec:experiments}).

This setup implies that in order to translate, we must apply UTD on both the training and test audio. Using additional (not only training) audio in UTD increases the likelihood of discovering more clusters. We therefore generate pseudotext for the combined audio, train the MT model on the pseudotext of the training audio, and apply it to the pseudotext of the test data. This is fair since the UTD has access to only the audio.\footnote{This is the simplest approach for our proof-of-concept system. In a more realistic setup, we could use the training audio to construct a consensus representation of each pseudoterm
\cite{petitjean2011global,antonios+chiang+duong_EMNLP2016},
then use DTW to identify its occurrences in test data to translate.}

\section{Dataset}
\label{sec:dataset}

Although we did not have access to a low-resource dataset, there is a corpus of noisy multi-speaker speech that simulates many of the conditions we expect to find in our motivating applications: the CALLHOME Spanish--English speech translation dataset (LDC2014T23; Post el al., 2013)\nocite{post2013improved}.\footnote{We did not use the Fisher portion of the corpus.}
We ran UTD over all 104 telephone calls, which pair $11$ hours of audio with Spanish transcripts  and their crowdsourced English translations. The transcripts contain 168,195 Spanish word tokens (10,674 types), and the translations contain 159,777 English word tokens (6,723 types).
Though our system does not require Spanish transcripts, we use them to evaluate UTD and to simulate a perfect UTD system, called the \textit{oracle}.

For MT training, we use the pseudotext and translations of 50 calls, and we filter out stopwords in the translations with NLTK \cite{bird2009natural}.\footnote{\texttt{http://www.nltk.org/}}
Since UTD is better at matching patterns from the same speaker (\textsection\ref{sub:splitting_words}), we created two types of 90/10\% train/test split: at the {\it call level} and at the {\it utterance level}.
For the latter, 90\% of the utterances are randomly chosen for the training set (independent of which call they occur in), and the rest go in the test set.
Hence at the utterance level, but not the call level, some speakers are included in both training and test data. Although the utterance-level split is optimistic, it allows us to investigate how multiple speakers affect system performance. In either case, the oracle has about 38k Spanish tokens to train on.

\section{Analysis of challenges from UTD}
\label{sec:utd_analysis}

Our system relies on the pseudotext produced by ZRTools (the only freely available UTD system we are aware of), which presents several challenges for MT. We used the default ZRTools parameters, and it might be possible to tune them to our task, but we leave this to future work.


\subsection{Assigning wrong words to a cluster}

Since UTD is unsupervised, the  discovered clusters are noisy.
Fig.~\ref{fig:utd_out} shows an example of an incorrect match between the acoustically similar~``{qué tal vas con}'' and ``{te trabajo y''} in utterances B and C, leading to a common assignment to c2.
Such inconsistencies in turn affect the translation distribution conditioned on {c2}.

Many of these
errors are due to cross-speaker matches, which are known to be more challenging for UTD~\cite{carlin+2011+rapideval,kamper2015unsupervised,sameer+2016+icassp+utd}.
Most matches in our corpus are across calls, yet these are also the least accurate (Table~\ref{tab:mt_x_call_uttr}). Within-utterance matches, which are always from the same speaker, are the most reliable, but make up the smallest proportion of the discovered pairs. Within-call matches fall in between.
Overall, average cluster purity is only $34$\%, meaning that $66$\% of discovered patterns do not match the most frequent type in their cluster.

\subsection{Splitting words across different clusters}\label{sub:splitting_words}

Although most UTD matches are across speakers, recall of cross-speaker matches is lower than for same-speaker matches. As a result, the same word from different speakers often appears in multiple clusters, preventing the model from learning good translations. ZRTools discovers 15,089 clusters in our data, though there are only 10,674 word types. Only 1,614 of the clusters map one-to-one to a unique word type, while a many-to-one mapping of the rest covers only 1,819 gold types (leaving 7,241 gold types with no corresponding cluster).

Fragmentation of words across clusters renders pseudoterms impossible to translate when they appear only in test and not in training.
Table~\ref{tab:mt_oov_stats} shows that these {\it pseudotext out-of-vocabulary (OOV)} words are frequent, especially in the call-level split. This reflects differences in acoustic patterns of different speakers, but also in their vocabulary --- even the oracle OOV rate is higher in the call-level split.

\begin{table}[t]
\begin{center}
\begin{tabularx}{\linewidth}{lCCc}
\hline
 & {\bf utterance} & {\bf call} & {\bf corpus} \\
\hline
Matches & 2\% & 17\% & 81\% \\
Accuracy & 78\% & 53\% & 8\% \\
\hline
\end{tabularx}
\end{center}
\caption{UTD matches within utterances, within calls and within the corpus. Matches within an utterance or call are usually from the same speaker.}
\label{tab:mt_x_call_uttr}
\end{table}

\begin{table}[t]
\begin{center}
\begin{tabularx}{\linewidth}{lCc}
\hline
 & {\bf utterance split} & {\bf call split} \\
\hline
Oracle & 420 (10\%) & 719 (17\%) \\
Pseudotext & 601 (29\%) & 892 (44\%) \\
\hline
\end{tabularx}
\end{center}
\caption{Number (percent) of out-of-vocabulary (OOV) word tokens or pseudoterms in the test data for different experimental conditions. }
\label{tab:mt_oov_stats}
\end{table}

\subsection{UTD is sparse, giving low coverage}

UTD is most reliable on long and frequently-repeated patterns, so many spoken words are not represented in the pseudotext, as in Fig.~\ref{fig:utd_out}.
We found that the patterns discovered by ZRTools match only 28\% of the audio.
This low coverage reduces training data size,  affects alignment quality, and adversely affects  translation, which is only possible when pseudoterms are present.
For almost half the utterances, UTD fails to produce any pseudoterm at all.



\begin{table*}[t]
\begin{center}
\small
\begin{tabular}{cp{3cm}p{3cm}p{4cm}p{3cm}}
\hline
& \textbf{source text} & \textbf{gold translation} & \textbf{oracle translation} & \textbf{utd translation} \\
\hline
1 & cómo anda \noindent{{\color{lightgray} el}} plan escolar & \noindent{{\color{lightgray} how}} \noindent{{\color{lightgray} is}} \noindent{{\color{lightgray} the}} \uline{school} plan \uline{going} & things whoa mean plan school & \uline{school} \uline{going} \\
2 & dile \noindent{{\color{lightgray} que}} \noindent{{\color{lightgray} le}} mando saludos & tell \noindent{{\color{lightgray} him}} \noindent{{\color{lightgray} that}} \noindent{{\color{lightgray} i}} \uline{say} \uline{hi} & tell send best says & \uline{say} \uline{hi} \\
3 & \noindent{{\color{lightgray} sí}} \noindent{{\color{lightgray} con}} dos dientes menos & \uline{yeah} \noindent{{\color{lightgray} with}} two \uline{teeth} less & two teeth less least & denture \uline{yeah} \uline{teeth} \\
4 & \noindent{{\color{lightgray} o}} dejando \noindent{{\color{lightgray} o}} dejando dos días & \noindent{{\color{lightgray} or}} giving \noindent{{\color{lightgray} or}} giving \uline{two} \uline{days} & improves apart improves apart two days &\uline{two} \uline{days} \\
5 & ah \noindent{{\color{lightgray} ya}} okey veintitrés \noindent{{\color{lightgray} de}} noviembre \noindent{{\color{lightgray} no}} & ah yeah okay \uline{twenty} third \noindent{{\color{lightgray} of}} \uline{november} \noindent{{\color{lightgray} no}} & oh ah okay another three fourth november & \uline{twenty} \uline{november} \\

\hline
\end{tabular}
\end{center}
\caption{Source text (left) paired with translations by humans (gold), oracle, and UTD-based system. Underlined words appear in UTD and the corresponding human translations.}
\label{tab:good_bad_preds}
\end{table*}


\section{Speech translation experiments}\label{sec:experiments}

We evaluate our system by comparing its output to the English translations on the test data. Since it translates only a handful of words in each sentence, BLEU, which measures accuracy of word sequences, is an inappropriate measure of accuracy.\footnote{BLEU scores for supervised speech translation systems trained on our data can be found in \newcite{kumar+etal_icassp14}.}
Instead we compute precision and recall over the content words in the translation.
We allow the system to guess $K$ words per test pseudoterm, so for each utterance, we compute the number of correct predictions as $corr@K = | pred@K~\cap~gold |$, where $pred@K~$ is the multiset of words predicted using $K$ predictions per pseudoterm
and $gold$ is the multiset of content words in the reference translation. For utterances where the reference translation has no content words, we use stop words.\todo{ADL: why? seems inconsistent. [SB] Otherwise we lose the training data from that utterance. ADL: Now I'm confused. Training or test?}
The utterance-level scores are then used to compute corpus-level \mbox{Precision@$K$} and \mbox{Recall@$K$}.

\begin{figure}[t]
    \centering
    \centerline{\includegraphics[width=0.95\linewidth]{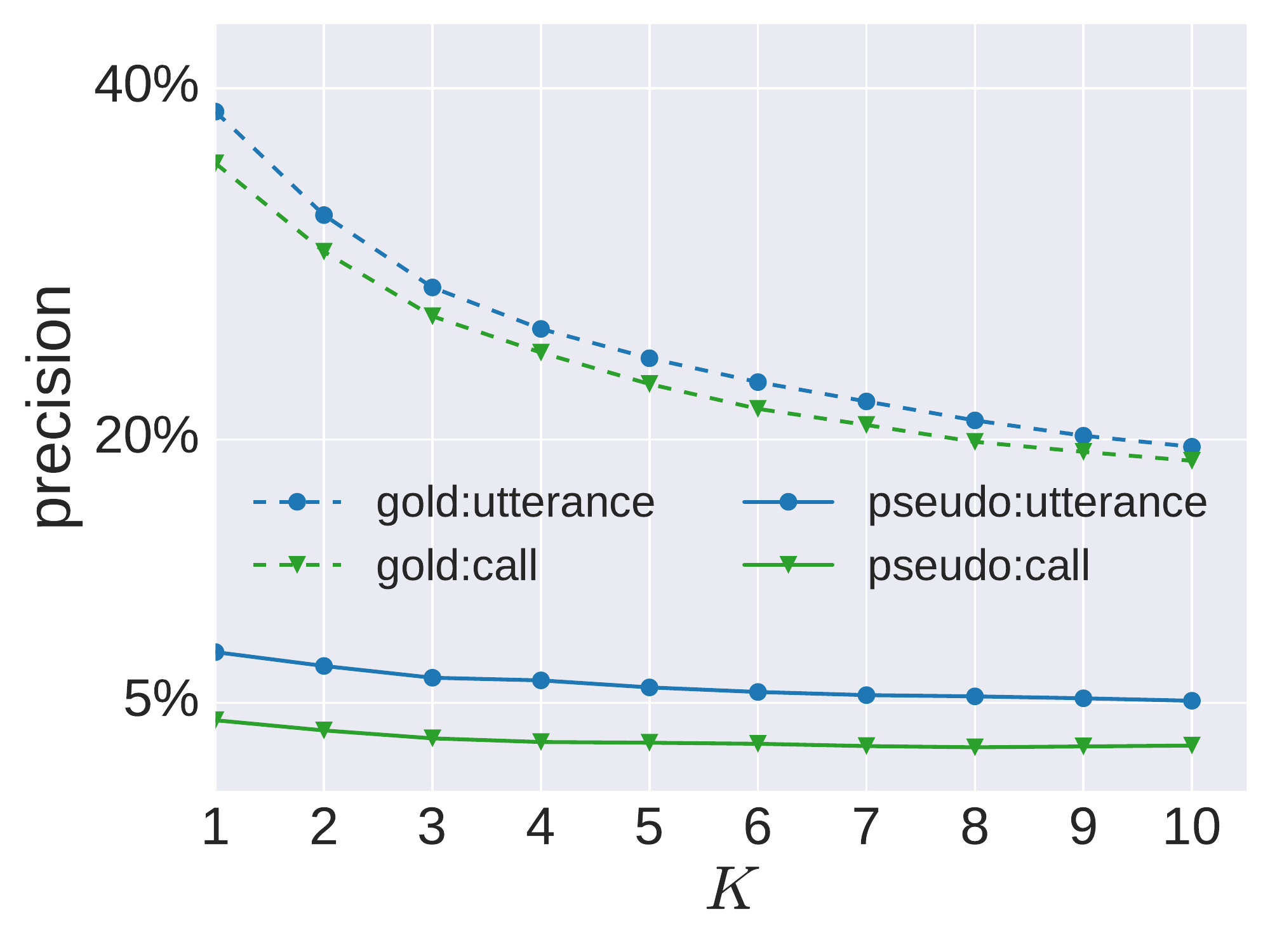}}
    \centerline{\includegraphics[width=0.95\linewidth]{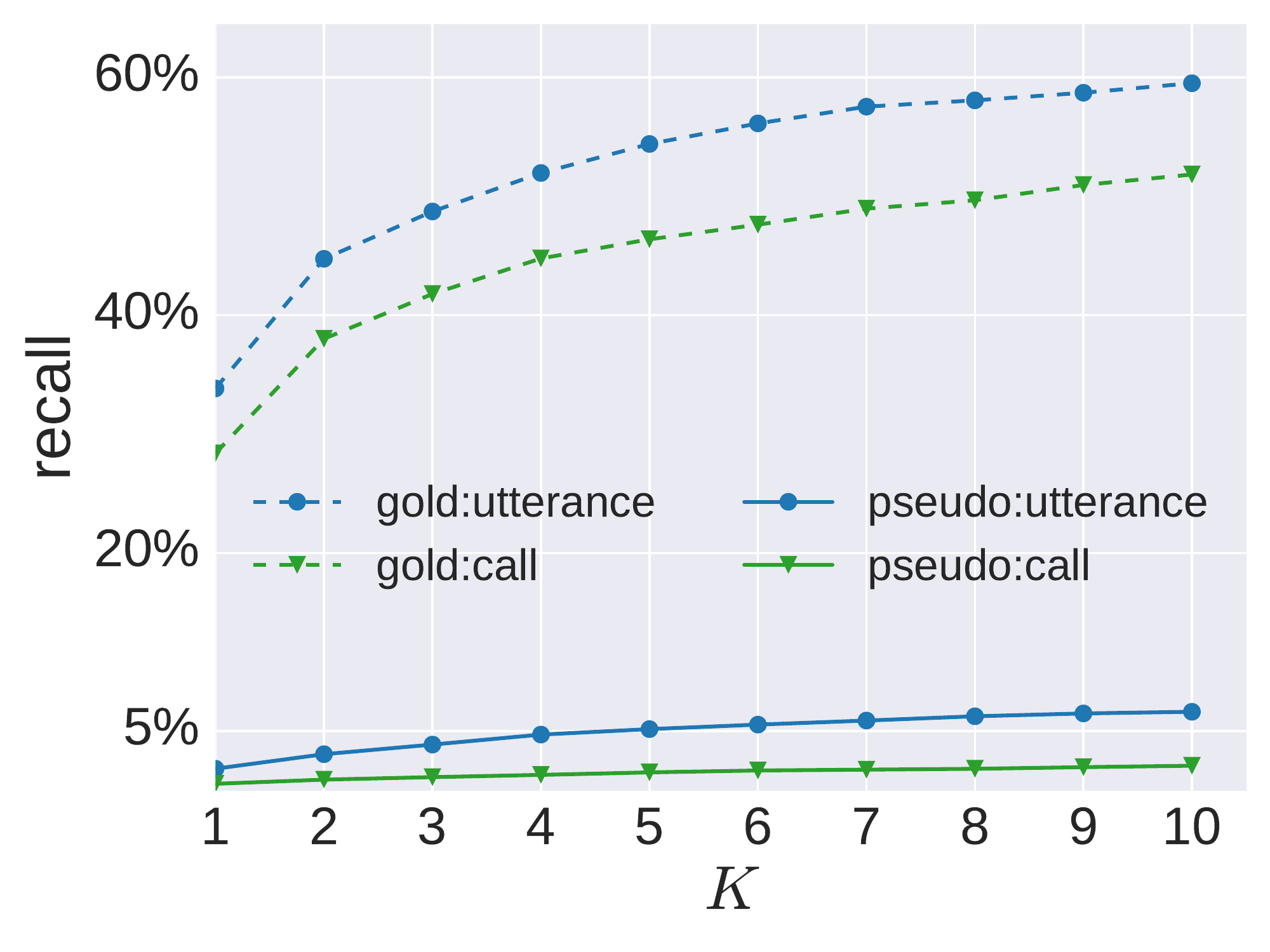}}
    \vspace*{-0.3\baselineskip}
    \caption{Precision and Recall $@K$ for the call and utterance level test sets.}
    \label{fig:p_r_dev}
    \vspace*{-0.7\baselineskip}
\end{figure}

Table~\ref{tab:mt_p_r_dev} and Fig.~\ref{fig:p_r_dev} show that even the oracle has mediocre precision and recall, indicating the difficulties of training an MT system using only bag-of-content-words on a relatively small corpus. Splitting the data by utterance works somewhat better, since training and test share more vocabulary.


\begin{table}[t]
\begin{center}
\begin{tabular}{cc|cc|cc}
\hline
& & \multicolumn{2}{c}{oracle} & \multicolumn{2}{|c}{pseudotext} \\
$K$ & metric & utterance & call & utterance & call \\
\hline
$1$ & Prec.& 38.6 & 35.7 & 7.9 & 4.0 \\
$1$ & Rec. & 33.8 & 28.4 & 1.8 & 0.6 \\
$5$ & Prec. & 24.6 & 23.1 & 5.9 & 2.7 \\
$5$ & Rec. & \bf{54.4} & 46.4 & \bf{5.2} & 1.5 \\
\hline
\end{tabular}
\end{center}
\caption{Precision and recall for $K=1$ and $K=5$ under different conditions.}
\label{tab:mt_p_r_dev}
\end{table}

Table~\ref{tab:mt_p_r_dev} and Fig.~\ref{fig:p_r_dev} also show a large gap between the oracle and our system. 
This is not surprising given the problems with the UTD output discussed in Section~\ref{sec:utd_analysis}. In fact, it is encouraging given the small number of discovered terms and the low cluster purity that our system can still correctly translate some words (Table~\ref{tab:good_bad_preds}). These results are a positive proof of concept, showing that it is possible to discover and translate keywords from audio data even with no ASR or MT system. Nevertheless, UTD quality is clearly a limitation, especially for the more realistic by-call data split.

\section{Conclusions and future work}

Our results show that it is possible to build a speech translation system using only source-language audio paired with target-language text, which may be useful in many situations where no other speech technology is available. Our analysis also points to several possible improvements. Poor cross-speaker matches and low audio coverage prevent our system from achieving a high recall, suggesting the of use speech features that are effective in multi-speaker settings~\cite{kamper2015unsupervised,kamper+etal_arxiv16} and speaker normalization \cite{zeghidour+etal_interspeech16}. Finally, \newcite{sameer+2016+icassp+utd} recently showed that UTD can be improved using the translations themselves as a source of information, which suggests joint learning as an attractive area for future work.

On the other hand, poor precision is most likely due to the simplicity of our MT model, and designing a model whose assumptions match our data conditions is an important direction for future work, which may combine our approach with insight from recent, quite different audio-to-translation models \cite{duong2015attentional,antonios+chiang+duong_EMNLP2016,adams+etal_interspeech16,adams+etal_emnlp16,berard+etal_nipsworkshop16}. Parameter-sharing using word and acoustic embeddings would allow us to make predictions for OOV pseudoterms by using the nearest in-vocabulary pseudoterm instead.






\section*{Acknowledgments}
\label{sec:thanks}
We thank David Chiang and Antonios Anastasopoulos for sharing alignments of the CALLHOME speech and transcripts; Aren Jansen for assistance with ZRTools; and Marco Damonte, Federico Fancellu, Sorcha Gilroy, Ida Szubert, Nikolay Bogoychev, Naomi Saphra, Joana Ribeiro and Clara Vania for comments on previous drafts. This work was supported in part by a James S McDonnell Foundation Scholar Award and a Google faculty research award.



\bibliography{eacl2017}

\begin{thebibliography}{}

\bibitem[\protect\citename{Adams \bgroup et al.\egroup
  }2016a]{adams+etal_interspeech16}
Oliver Adams, Graham Neubig, Trevor Cohn, and Steven Bird.
\newblock 2016a.
\newblock Learning a translation model from word lattices.
\newblock In { Proc. Interspeech}.

\bibitem[\protect\citename{Adams \bgroup et al.\egroup
  }2016b]{adams+etal_emnlp16}
Oliver Adams, Graham Neubig, Trevor Cohn, Steven Bird, Quoc~Truong Do, and
  Satoshi Nakamura.
\newblock 2016b.
\newblock Learning a lexicon and translation model from phoneme lattices.
\newblock In { Proc. EMNLP}.

\bibitem[\protect\citename{Anastasopoulos \bgroup et al.\egroup
  }2016]{antonios+chiang+duong_EMNLP2016}
Antonios Anastasopoulos, David Chiang, and Long Duong.
\newblock 2016.
\newblock An unsupervised probability model for speech-to-translation alignment
  of low-resource languages.
\newblock In { Proc. EMNLP}.

\bibitem[\protect\citename{Bansal \bgroup et al.\egroup
  }2017]{sameer+2016+icassp+utd}
Sameer Bansal, Herman Kamper, Sharon Goldwater, and Adam Lopez.
\newblock 2017.
\newblock Weakly supervised spoken term discovery using cross-lingual side
  information.
\newblock In { Proc. ICASSP}.

\bibitem[\protect\citename{Berard \bgroup et al.\egroup
  }2016]{berard+etal_nipsworkshop16}
Alexandre Berard, Olivier Pietquin, Christophe Servan, and Laurent Besacier.
\newblock 2016.
\newblock Listen and translate: A proof of concept for end-to-end
  speech-to-text translation.
\newblock In { {NIPS} Workshop on End-to-end Learning for Speech and Audio
  Processing}.

\bibitem[\protect\citename{Besacier \bgroup et al.\egroup
  }2006]{besacier+etal_slt06}
Laurent Besacier, Bowen Zhou, and Yuqing Gao.
\newblock 2006.
\newblock Towards speech translation of non written languages.
\newblock In { Proc. SLT}.

\bibitem[\protect\citename{Besacier \bgroup et al.\egroup }2014]{besacier2014}
Laurent Besacier, Etienne Barnard, Alexey Karpov, and Tanja Schultz.
\newblock 2014.
\newblock Automatic speech recognition for under-resourced languages: A survey.
\newblock { Speech Communication}, 56:85--100.

\bibitem[\protect\citename{Bird \bgroup et al.\egroup }2009]{bird2009natural}
Steven Bird, Ewan Klein, and Edward Loper.
\newblock 2009.
\newblock { Natural language processing with Python}.
\newblock O'Reilly Media.

\bibitem[\protect\citename{Brown \bgroup et al.\egroup }1993]{brown+1993}
Peter~F Brown, Vincent J~Della Pietra, Stephen A~Della Pietra, and Robert~L
  Mercer.
\newblock 1993.
\newblock The mathematics of statistical machine translation: Parameter
  estimation.
\newblock { Computational Linguistics}, 19(2):263--311.

\bibitem[\protect\citename{Carlin \bgroup et al.\egroup
  }2011]{carlin+2011+rapideval}
Michael~A Carlin, Samuel Thomas, Aren Jansen, and Hynek Hermansky.
\newblock 2011.
\newblock Rapid evaluation of speech representations for spoken term discovery.
\newblock In { Proc. Interspeech}.

\bibitem[\protect\citename{Duong \bgroup et al.\egroup
  }2016]{duong2015attentional}
Long Duong, Antonios Anastasopoulos, David Chiang, Steven Bird, and Trevor
  Cohn.
\newblock 2016.
\newblock An attentional model for speech translation without transcription.
\newblock In { Proc. NAACL HLT}.

\bibitem[\protect\citename{Dyer \bgroup et al.\egroup
  }2013]{dyer+2013+fastalign}
Chris Dyer, Victor Chahuneau, and Noah~A Smith.
\newblock 2013.
\newblock A simple, fast, and effective reparameterization of {IBM} model 2.
\newblock In { Proc. ACL}.

\bibitem[\protect\citename{Jansen and Van~Durme}2011]{jansen2011efficient}
Aren Jansen and Benjamin Van~Durme.
\newblock 2011.
\newblock Efficient spoken term discovery using randomized algorithms.
\newblock In { Proc. ASRU}.

\bibitem[\protect\citename{Kamper \bgroup et al.\egroup
  }2015]{kamper2015unsupervised}
Herman Kamper, Micha Elsner, Aren Jansen, and Sharon Goldwater.
\newblock 2015.
\newblock Unsupervised neural network based feature extraction using weak
  top-down constraints.
\newblock In { Proc. ICASSP}.

\bibitem[\protect\citename{Kamper \bgroup et al.\egroup
  }2016a]{kamper+etal_arxiv16}
Herman Kamper, Aren Jansen, and Sharon Goldwater.
\newblock 2016a.
\newblock A segmental framework for fully-unsupervised large-vocabulary speech
  recognition.
\newblock { arXiv preprint arXiv:1606.06950}.

\bibitem[\protect\citename{Kamper \bgroup et al.\egroup
  }2016b]{kamper+etal_taslp16}
Herman Kamper, Aren Jansen, and Sharon Goldwater.
\newblock 2016b.
\newblock Unsupervised word segmentation and lexicon discovery using acoustic
  word embeddings.
\newblock { IEEE/ACM Trans. Audio, Speech, Language Process.}, 24(4):669--679.

\bibitem[\protect\citename{Kumar \bgroup et al.\egroup
  }2014]{kumar+etal_icassp14}
Gaurav Kumar, Matt Post, Daniel Povey, and Sanjeev Khudanpur.
\newblock 2014.
\newblock Some insights from translating conversational telephone speech.
\newblock In { Proc. ICASSP}.

\bibitem[\protect\citename{Lee \bgroup et al.\egroup }2015]{lee+etal_tacl15}
Chia-ying Lee, T~O'Donnell, and James Glass.
\newblock 2015.
\newblock Unsupervised lexicon discovery from acoustic input.
\newblock { Trans. ACL}, 3:389--403.

\bibitem[\protect\citename{Martin \bgroup et al.\egroup }2015]{AlanBlackStuff}
Lara~J Martin, Andrew Wilkinson, Sai~Sumanth Miryala, Vivian Robison, and
  Alan~W Black.
\newblock 2015.
\newblock Utterance classification in speech-to-speech translation for
  zero-resource languages in the hospital administration domain.
\newblock In { Proc. ASRU}.

\bibitem[\protect\citename{Munro}2010]{munro2010}
Robert Munro.
\newblock 2010.
\newblock Crowdsourced translation for emergency response in {Haiti}: the
  global collaboration of local knowledge.
\newblock In { AMTA Workshop on Collaborative Crowdsourcing for Translation}.

\bibitem[\protect\citename{Park and Glass}2008]{park2008unsupervised}
Alex~S Park and James Glass.
\newblock 2008.
\newblock Unsupervised pattern discovery in speech.
\newblock { IEEE Trans. Audio, Speech, Language Process.}, 16(1):186--197.

\bibitem[\protect\citename{Petitjean \bgroup et al.\egroup
  }2011]{petitjean2011global}
Fran{\c{c}}ois Petitjean, Alain Ketterlin, and Pierre Gan{\c{c}}arski.
\newblock 2011.
\newblock A global averaging method for dynamic time warping, with applications
  to clustering.
\newblock { Pattern Recognition}, 44(3):678--693.

\bibitem[\protect\citename{Post \bgroup et al.\egroup }2013]{post2013improved}
Matt Post, Gaurav Kumar, Adam Lopez, Damianos Karakos, Chris Callison-Burch,
  and Sanjeev Khudanpur.
\newblock 2013.
\newblock Improved speech-to-text translation with the {F}isher and {C}allhome
  {S}panish--{E}nglish speech translation corpus.
\newblock In { Proc. IWSLT}.

\bibitem[\protect\citename{Versteegh \bgroup et al.\egroup
  }2015]{versteegh+etal_interspeech15}
Maarten Versteegh, Roland Thiolli{\`e}re, Thomas Schatz, Xuan~Nga Cao, Xavier
  Anguera, Aren Jansen, and Emmanuel Dupoux.
\newblock 2015.
\newblock The {Zero Resource Speech Challenge} 2015.
\newblock In { Proc. Interspeech}.

\bibitem[\protect\citename{Waibel and Fugen}2008]{waibel+fugen_ieee08}
Alex Waibel and Christian Fugen.
\newblock 2008.
\newblock Spoken language translation.
\newblock { IEEE Signal Processing Magazine}, 3(25):70--79.

\bibitem[\protect\citename{Zeghidour \bgroup et al.\egroup
  }2016]{zeghidour+etal_interspeech16}
Neil Zeghidour, Gabriel Synnaeve, Nicolas Usunier, and Emmanuel Dupoux.
\newblock 2016.
\newblock Joint learning of speaker and phonetic similarities with {Siamese}
  networks.
\newblock In { Proc. Interspeech}.

\end{thebibliography}
\bibliographystyle{eacl2017}

\end{document}